\documentclass{article}
\usepackage{amsmath,amsthm,amssymb}
\usepackage{amsfonts}
\usepackage[margin=1in]{geometry}
\usepackage{graphicx}
\usepackage{subcaption}
\usepackage{bm}

\usepackage{authblk}

\usepackage{listings}
\newtheorem{example}{Example}
  
\title{Solving Boltzmann Optimization Problems with Deep Learning}
 
\author[*]{Fiona M. Knoll}
\author[**]{John T. Daly}
\author[**]{Jess J. Meyer}
\affil[*]{Department of Computer Science, US Naval Academy, Annapolis, MD 21402}
\affil[**]{Advanced Computing Systems, Laboratory for Physical Sciences, Catonsville, MD 21228 }

\date{}

\begin{document}
 
\maketitle

\begin{abstract}
Decades of exponential scaling in high performance computing (HPC) efficiency is coming to an end. Transistor based logic in complementary metal-oxide semiconductor (CMOS) technology is approaching physical limits beyond which further miniaturization will be impossible. Future HPC efficiency gains will necessarily rely on new technologies and paradigms of compute. The Ising model shows particular promise as a future framework for highly energy efficient computation. Ising systems are able to operate at energies approaching thermodynamic limits for energy consumption of computation. Ising systems can function as both logic and memory. Thus, they have the potential to significantly reduce energy costs inherent to CMOS computing by eliminating costly data movement. The challenge in creating Ising-based hardware is in optimizing useful circuits that produce correct results on fundamentally nondeterministic hardware. The contribution of this paper is a novel machine learning approach, a combination of deep neural networks and random forests, for efficiently solving optimization problems that minimize sources of error in the Ising model. In addition, we provide a process to express a Boltzmann probability optimization problem as a supervised machine learning problem. 
\end{abstract}

%%%%%%%%%%%%%%%%%%% Introduction %%%%%%%%%%%%%%%%%%%
Since the creation of the first microprocessor in 1971, the number of transistors on a chip has grown at an exponential rate that has only recently begun to abate~\cite{theis2017}. This trend was first observed in 1975 by engineer Gordon Moore~\cite{moore1975}, and has been known as Moore's law ever since. Today there is a growing consensus that Moore's law is coming to an end.  
Growth in the number of transistors is reaching physical limits, processing frequency has largely stagnated, and power consumption has effectively plateaued. The usability of systems based on CMOS devices creates additional challenges: deeply scaled technologies exhibiting unexplained aging effects, greater hardware variability, increased soft error susceptibility, and decreased noise immunity because of operating at near-threshold voltages~\cite{seifert2012}. All told, the path forward for continued improvement in HPC performance and efficiency must consider architectures beyond von Neumann and technologies beyond CMOS, where nondeterminism is not an accident but ``an essential part of the process under consideration.''~\cite{vonNeumann1952} Along these lines, Ising-based hardware has shown particular promise as an extremely energy efficient approach to solving particular classes of hard problems~\cite{cai2023unconventional}, \cite{haribara2017performance}, \cite{hong2016}, \cite{huckaba2022}, \cite{aadit2022}, \cite{yamamoto2017coherent}.

The architecture of Ising-based hardware arises from the Ising model, a famous model from statistical mechanics. The model is a lattice structure, where each site on the (potentially infinite) lattice is referred to as a \emph{spin}. Unlike the quantum world the spins are discrete, taking only two values $\{-1, +1\}$. The spins are interdependent and the probability of the spins taking on a set of positions (known as the state) can be measured by the Boltzmann distribution, also known as the Gibb's distribution. 

By increasing the spin space of the system such that each spin can take values in $\{1, \dots, q\}$ for some integer $q \geq 2$, we obtain the Potts model \cite{wu1982potts}. 
Note that for $q=2$, the Potts model is the Ising model.
The Potts model is applied across multiple disciplines. For example, in statistical mechanics and computational biology fields, the model is used to study various macroscopic properties of the (biological or physical) system such as phase transitions, entropy, and thermodynamic properties~\cite{baxter2016exactly}, \cite{broderick2007faster}, \cite{dubois2009boltzmann}.  
For more complete surveys of various applications of the Potts model, see  \cite{nguyen2017inverse}, \cite{rozikov2103gibbs}.

For $q=2$, the Potts / Ising model has been used to design non-von Neumann logical circuits, which is the application motivating this work~\cite{martinmoore}. 
These circuits are inherently nondeterministic: for a given logic operation, every possible output has some probability of occurring.  
To design an Ising-based circuit, we define an optimization problem of the Boltzmann probability of the system. The objective is to choose system parameters such that the configurations corresponding to correct outputs have high probability. 

An obstacle for working with the Ising model is the computational complexity involved in calculating the Boltzmann distribution and its requisite partition function. This is computationally intractable for even simple problems. Consequently, various approximations to compute this distribution efficiently are of continued research interest in many disciplines, e.g.,~\cite{ramupartition}, \cite{haddadanfast}, \cite{PhysRevA.80.022340}.
Our approach bypasses this challenge. We craft our problem as an optimization problem that is tractable and possible to solve with standard solvers. We successfully train deep machine learning models to solve this optimization problem. 

In Section~\ref{background} we introduce the problem of solving for the parameters of the Ising system, known as the reverse Ising problem, and motivate the associated Boltzmann probability optimization problem. In Section~\ref{modeling} we detail the process of converting this optimization problem to one amenable to efficient data generation required for supervised machine learning, and describe our machine learning models. In Section~\ref{comparisons} we analyze the performance improvements of our models over traditional optimization methods. The contribution of this work is a novel approach to solving the otherwise intractable problem of optimizing Ising system parameters to produce desired outputs from specified inputs with high probability.  
Our work provides a method of efficiently predicting the dynamics of Ising systems, enabling the tackling of more complex problems on this exciting next-gen class of hardware.
%The contributions of this paper are twofold. First, we demonstrate a means of expressing the Boltzmann probability that can be computed efficiently for the reverse Ising problem. Second, we show that our approach can be used to generate effective training data that can be modeled. 

 %%%%%%%%%%%%%%%%%%% Background %%%%%%%%%%%%%%%%%%%

\section{Background of the Boltzmann Optimization Problem}
\label{background} 
    
\subsection{The Reverse Ising Problem} \label{RIModel}
    The \emph{Ising Model} represents a system of spins. In the classical setting, these spins take on the values $\pm 1$ and can be representative of bit values 0 and 1. The \emph{state} of the system is defined as the set of spin values.
  
    As with magnets, neighboring spins interact with one another and have a \emph{ground state}, the state to which the spins naturally settle.
    In the Ising model, the ground state is that with the lowest energy, which is analogous to magnets aligning to their ground
    state due to the interaction of their magnetic fields.
    Since the Ising model is a nondeterministic model derived from statistical mechanics, the probability of settling to the ground state is a function of that state's energy. The resulting probability distribution is discussed further in Section \ref{SectionBoltz}.
    
    The total energy, both kinetic and potential, of an 
    Ising model with state $\textbf{s}=\{s_i\} \in \{\pm1\}^N$ is defined by the \emph{Hamiltonian} of the
    state:
    \[H(\textbf{s}) = \sum_{i=1}^N h_i \cdot s_i + 
            \sum_{\langle i,j\rangle} J_{i,j}\cdot s_i \cdot s_j,\]   
    where $h_i$, $J_{i,j} \in \mathbb R$ are the energy fields affecting individual spins and the coefficients representing the interactions between
    the spins, respectively, and
    the second sum is over pairs of neighboring spins (every pair is counted once). 
    
    %%%%%%%%%%%%%%%%%%%% The Classic Ising Problem %%%%%%%%%%%%%%%%%%%% 
    The classic Ising problem is as follows: given coefficients $h_i$, $J_{i,j} \in \mathbb R$,
    find state $\textbf{s} \in \{\pm 1\}^N$ where $\textbf{s}$ has the minimum Hamiltonian value. The state
    with the smallest Hamiltonian value, i.e., the lowest energy, will be the
    resulting ground state with higher probability than any other possible state of the system. 
    This problem has been shown to be NP-complete \cite{IMComplete}.

    %%%%%%%%%%%%%%%%%%%% The Reverse Ising Problem%%%%%%%%%%%%%%%%%%%%     
    The \emph{reverse Ising problem} takes a slight spin off the classic.
    Suppose we have a system with $N$ spins and the spins are partitioned 
    into two sets of sizes $n$ and $m$, that is, $N=n+m$.
    Without loss of generality, suppose the first $n$ spins $(s_1,\ldots,s_n)$ are fixed, 
    i.e. the values of the spins are known
    and programmed into the system. And, suppose the last $m$ spins are not fixed, that is not
    programmed into the system, but they however have preferred known values: $(s_{n+1}, ..., s_N)$.
    The question then is, can we find Hamiltonian coefficients such that the
    desired state $\textbf{s}=(s_1, \ldots, s_n, s_{n+1}, \ldots, s_N)$ will become the ground
    state with high probability, that is $\textbf{s}$ has the smallest total energy of all states given
    Hamiltonian coefficients $h_i$, $J_{i,j}$.  
    More specifically, given state $\textbf{s}=(s_1, \ldots, s_n, s_{n+1}, \ldots, s_N)
    \in \{\pm 1\}^N$ find coefficients $h_i$, $J_{i,j} \in \mathbb R$ such that
        \[H(s_1, \ldots, s_n, s_{n+1}, \ldots, s_N) < H(s_1, \ldots, s_n, t_1, \ldots, t_m)\]
    for $t_k \in \{\pm 1\}$, $1 \leq k \leq m$, where $t_k\neq s_{n+k}$ for at least one $k$. 
    In addition, since Ising systems are nondeterministic, we maximize the event of state $\textbf{s}$ occurring by minimizing
    the energy of that state: 
        \[\min_{\psi} H_\psi (\textbf{s}),\]
    where 
            \[\psi = \{h_1, h_2,\ldots, h_N, J_{1,2}, \ldots, J_{N-1,N}\}\] 
    is the vector representing the
    Hamiltonian coefficients of a system with $N$ spins and $H_\psi$ the Hamiltonian of a state $\textbf{s}$ with 
    Hamiltonian coefficients $\psi$.

    \begin{example} \label{example_opt}
         Let $\textbf{s}=(-1,1,-1)$ where the third spin
         can vary. We then desire to find Hamiltonian coefficients 
        that will ensure the state $\textbf{s}=(-1,1,-1)$ will have the smallest energy
        of the states $(-1,1,-1)$ and $(-1,1,1)$:
        \begin{align*} &\min_{\psi} H_\psi(-1,1,-1) \\%H_\psi(\textbf{s}) \\
            &\text{s.t. } H_\psi(-1,1,-1) < H_\psi(-1,1,1) \end{align*}
    \end{example}

    We generalize Example \ref{example_opt} to the higher dimension $N=n+m$, 
    and obtain the following optimization problem: \\
    Let $\textbf{s}= (\textbf{u}, \textbf{v}) \in \{\pm1\}^N$ be the desired state, where $\textbf{u}\in\{\pm1\}^n$ and $\textbf{v}\in \{\pm1\}^m$. 
    Then, the problem of interest becomes
        \begin{align} 
            \min_{\psi} \; & H_\psi(\textbf{s}= (\textbf{u}, \textbf{v})) \label{eq_H0}\\
            &\text{s.t. } H_\psi(\textbf{u}, \textbf{v}) < H_\psi(\textbf{u}, \textbf{t}) \text{ for all $\textbf{t} \neq \textbf{v}$.} \nonumber
        \end{align}

    However, the system of inequalities $\{H_\psi(\textbf{u},\textbf{v}) < H_\psi(\textbf{u},\textbf{t})\}_{\textbf{t} \neq \textbf{v}}$ is not feasible in certain scenarios. Consequently, in such cases extra degrees of freedom are introduced by adding more spins, referred to as \emph{auxiliary spins}. Let $\alpha$ represent the number of auxiliary spins added to the system and let $\textbf{a} \in \{\pm1\}^\alpha$
    represent the vector of auxiliary spins. From now on, we consider the number of spins to be partitioned as $N=n+m+\alpha$.  For each state of desired spins $s_1, \ldots, s_{n+m}$ we want to find the position of auxiliary spins that give us the lowest energy $H(\textbf{s})$. 
    Hence, we minimize over the auxiliary spins in addition to the Hamiltonian 
    coefficients, and equation (\ref{eq_H0}) becomes 
        \begin{align} 
            \min_{\substack{ \psi \\ \textbf{a} \in \{\pm1\}^\alpha}} & H_\psi(\textbf{s}= (\textbf{u},\textbf{v},\textbf{a})) \label{eq_H1} \\
            &\text{s.t. for all $\textbf{t} \neq \textbf{v}$} \nonumber \\
            & \quad \quad H_\psi(\textbf{u}, \textbf{v}, \textbf{a}) < H_\psi(\textbf{u}, \textbf{t}, \textbf{a}) \nonumber\\
            &\quad \quad H_\psi (\textbf{u},\textbf{v},\textbf{a}) < H_\psi (\textbf{u},\textbf{t},-a_1, a_2, \ldots, a_\alpha) \nonumber \\
            &\quad \quad H_\psi (\textbf{u},\textbf{v},\textbf{a}) < H_\psi (\textbf{u},\textbf{t}, a_1, -a_2, \ldots, a_\alpha) \nonumber \\
            & \quad \quad \quad   \vdots \nonumber \\
            &\quad \quad H_\psi (\textbf{u},\textbf{v},\textbf{a}) < H_\psi (\textbf{u},\textbf{t},-a_1, -a_2, \ldots, -a_\alpha) \nonumber \;.
        \end{align}

    In the reverse Ising problem, the above is generalized to a set of desired states:
    For a set of states $S = \{\textbf{s}^{(i)}\}$,
    $1 \leq i \leq \ell$, find Hamiltonian coefficients $\psi = \{h_1, \ldots, h_N, J_{1,2}, \ldots, J_{N-1,N}\}$ 
    such that for each state $\textbf{s}^{(i)} = (s_1^{(i)}, \ldots, s_N^{(i)})$ the total energy of the
    system is smaller than that of all possible sets of spins with the $n$ spins being
    the same. That is,
        for $1\leq i \leq \ell$,
        \[H_\psi(s_1^{(i)}, \ldots, s_n^{(i)}, s_{n+1}^{(i)}, \ldots, s_{n+m}^{(i)}, a_1^{(i)}, \ldots, a_\alpha^{(i)}))
            < H_\psi(s_1^{(i)}, \ldots, s_n^{(i)}, t_1, \ldots, t_m, a_1^{(i)}, \ldots, a_\alpha^{(i)})\]
    for $t_k \in \{\pm 1\}$ where $t_k \neq s_{n+k}^{(i)}$ for at least one value of $k$. 
    We note that for each desired state $\textbf{s}= (s_1, \ldots, s_{n+m})$, the auxiliary spins $a_1, \ldots, a_\alpha$ can take on different spin values. For example, if there are $3$ desired states, then there are $3$ auxiliary spin vectors $\textbf{a}$ that are completely independent of one another.

   Recall that the Ising model is a nondeterministic system where the probability of settling to a state, including that with lowest energy, is a function of that state's energy. As a result, rather than minimizing the energy of a set of states, we turn to the probability of the states.

\subsection{The Boltzmann Probability Objective Function} \label{SectionBoltz}
    In 1868, Ludwig Boltzmann discovered a probability distribution, the Boltzmann distribution, that models the likelihood of states occurring within a given system~\cite{boltzmann2015relationship}.
     The probability of the system taking on state $\tilde{s}$ given Hamiltonian coefficients $\psi$ is defined by the \emph{Boltzmann probability}:
    \begin{equation} \label{boltz}
        \text{Prob} (S = \tilde{s} ) := \frac{ \exp(H_\psi(\tilde{s})/ (\beta T) ) }{\sum_{s \in \{\pm 1\}^N} \exp(H_\psi(s)/ (\beta T) )},
    \end{equation}
    where $\beta$ represents the Boltzmann constant and $T$ the temperature.
    In this paper, we consider both the Boltzmann constant $\beta$ and temperature to be fixed. More specifically, we let $\beta T = 1$.

    Viewing the probability of a system taking on state $\tilde{s}$ in terms of the probability of all other states, equation 
    (\ref{boltz}) is equivalent to
        \begin{align*}
        1 - \sum_{w \neq \tilde{s}} \text{Prob} (S = w)
            & = 1 - \sum_{w \neq \tilde{s}} \frac{\exp(H_\psi(w))}{\sum_s \exp(H_\psi(s))} \\ 
            & = 1 - \frac{\sum_{w \neq \tilde{s}} \exp (H_\psi(w)) } {\exp(H_\psi(\tilde{s})) + \sum_{w\neq \tilde{s}} \exp (H_\psi(w)) }.
        \end{align*}

    In light of the reverse Ising problem, we desire to maximize the probability of
    obtaining a set of desired states, which we will notate as 
    $\{ \textbf{s}^{(i)} = (\textbf{u}^{(i)}, \textbf{v}^{(i)}, \textbf{a}^{(i)}) \}$, where 
    $\textbf{a}$ represents the auxiliary spins that are free to take on any value. 
    Equivalently, we want to minimize the maximum probability of obtaining a state with an incorrect second
    portion $\textbf{v}$ over all the possible first portions $\textbf{u}$.
    
     More specifically,
    let $S = \{ \textbf{s}^{(i)} = (\textbf{u}^{(i)}, \textbf{v}^{(i)}, \textbf{a}^{(i)}) \}_{1\leq i\leq \ell}$ represent the set of states desired, where $\textbf{a}$ is free.
    Let $W^{(i)} = \{ (\textbf{u}^{(i)}, \textbf{t}, \textbf{a}^{(i)}) \}_{\textbf{t}} $, where $\textbf{t}$ ranges over all the possible vectors in $\{\pm 1\}^m$ such that $\textbf{t} \neq \textbf{v}^{(i)}$. That is, $W^{(i)}$ represents the set of states where the second portion differs from $\textbf{v}^{(i)}$ in at least one position. Then, the probability of not being in state $\textbf{s}^{(i)} = (\textbf{u}^{(i)}, \textbf{v}^{(i)}, \textbf{a}^{(i)}) $ but rather in a state in $W^{(i)}$ is
\begin{align*}
    \text{Prob}_{boltz} (\textbf{s}^{(i)}, W^{(i)}) &:= \sum_{w\in W^{(i)}}\text{Prob} (S =w ) \\
    &= \frac{\sum_{w\in W^{(i)}} \exp (H_\psi(w)) } {\exp(H_\psi(\textbf{s}^{(i)})) + \sum_{w\in W^{(i)}} \exp (H_\psi(w)) } 
\end{align*}
    
    The resulting objective function of interest, which will be used through the rest of 
    the paper, is as follows: 
        \begin{align} \label{eq_opt_a}
        \rho(\textbf{a}^{(1)}, \ldots, \textbf{a}^{(\ell)}) &= 
        1 - \min_\psi \max_{1 \leq i \leq \ell}\ \text{Prob}_{boltz} (\textbf{s}^{(i)}, W^{(i)}) \nonumber\\
        &= 1 - \min_{\psi} 
            \max_{1 \leq i \leq \ell}\ \frac{\sum_{w\in W^{(i)}} \exp (H_\psi(w)) } {\exp(H_\psi(s^{(i)})) + \sum_{w\in W^{(i)}} \exp (H_\psi(w)) } ,
        \end{align}
    where $\psi$ is the set of Hamiltonian coefficients and $\textbf{a}^{(i)} \in \{\pm 1\}^\alpha$ for $1\leq i \leq \ell$, each $i$ corresponding to
    the desired state $\textbf{s}^{(i)}$.
    Throughout the rest of the paper, $(\textbf{a}^{(1)}, \ldots, \textbf{a}^{(\ell)})$ is referred to as the \emph{auxiliary array}.

\subsection{Size of the System and Complexity of the Problem}
    The size and complexity of the reverse Ising problem is determined by the number of fixed, non-fixed, and auxiliary spins of the system under consideration.
    For $N=n+m+\alpha$ total spins, the total number of possible states considered when computing the Boltzmann probability is $2^{N}$, and the vector $\psi$ of Hamiltonian coefficients has $\frac{1}{2}(N^2-N+2)$ variables. The number of linear inequality constraints in our resulting optimization problem is $2^{n+\alpha}(2^n-1)$.
    Finally the number of possible combinations of auxiliary spin values scales double exponentially as $2^{\alpha 2^n}$ since we consider $2^n$ desired states and auxiliary spins are free.
    So, where a system with 6 total spins (2 auxiliary spins) will have 48 inequality constraints and 256 possible auxiliary arrays, a problem with twice as many total spins (4 auxiliary spins) will have 3,840 inequality constraints and $2^{64} \approx 1.8 \times 10^{19}$ possible auxiliary arrays. 

    If the system had no auxiliary spins, the constraints are linear in $\psi$ since the Hamiltonian is linear in $\psi$. With the addition of auxiliary spins, the constraints become non-linear and non-convex. Further, the objective function (equation (\ref{eq_opt_a})) is a nonlinear, non-differentiable function with an intractable number of local minima. This rules out the possibility of using standard simplex methods to find a global minimum.
    Because of this, faster methods for evaluating the Boltzmann probability for arbitrary sized systems with large numbers of auxiliary spins are essential to optimization problems like the reverse Ising problem.

 %%%%%%%%%%%%%%%%%%% ModelConstruction %%%%%%%%%%%%%%%%%%%

\section{Modeling the Boltzmann Probability Optimization Problem}\label{modeling}

    We continue to assume there are a total of $N = n + m + \alpha$ spins in the system,
    where $\alpha$ is the number of auxiliary spins, $n$ spins are fixed, leaving $m$ of the spins to vary.
 Recall our minimization maximization programming problem from Section \ref{SectionBoltz}:
        \begin{align} \label{eq_opt_b}
            \rho(\textbf{a}^{(1)}, \ldots, \textbf{a}^{(\ell)}) = 1 - \min_{\psi} 
            \max_{1\leq i\leq \ell} \mathbf{\text{Prob}_{boltz}} \left(\textbf{s}^{(i)}, W^{(i)} \right) .
        \end{align}
where auxiliary spin values $\textbf{a}^{(i)} \in \{\pm 1\}^\alpha$ are the input, $\{ \textbf{s}^{(i)} = (\textbf{u}^{(i)}, \textbf{v}^{(i)}, \textbf{a}^{(i)}) \}$ represent the set of states desired and $W^{(i)} = \{ (\textbf{u}^{(i)}, \textbf{t}, \textbf{a}^{(i)}) \}_{\textbf{t} \neq \textbf{v}^{(i)}}$ represents the set of states that are the complement of state $\textbf{s}^{(i)}$.
    For a given problem size and a desired set of states, our goal is 
    to create a model that provides a means of calculating $\rho$ both quickly and 
    accurately, within some epsilon.

To do this, we first need to generate training data.
Equation (\ref{eq_opt_b}) presents several challenges for generating such data. Firstly it is numerically unstable, and secondly it is not everywhere-differentiable, ruling out the possibilities of using desirable gradient based methods. 
We describe the process via which we construct a numerically stable approximation to equation (\ref{eq_opt_b}). 
This development is singularly responsible for our ability to create high-quality training data required to train supervised machine learning models. 
This proved essential to making worthwhile progress towards our stated goal of optimizing Ising system parameters.
Following the description of this process, we describe the two machine learning models that we consider in this work. 

 \subsection{Generating the Training Data} \label{SectionData}
    %We desire to have a set of values $(\textbf{a}, \rho(\textbf{a}))$ specific to a problem size.
    
    We desire to transform equation (\ref{eq_opt_b}) such that we can easily and reliably compute solutions.
    The nonlinear optimization solver we used is based on the Sequential Least Squares Programming (SLSQP) algorithm, proposed by Dieter Kraft in 1988 in \cite{kraft1988software}. The solver is available through the Python open-source library SciPy.  

    For the majority of choices for Hamiltonian coefficients $\psi = (h_1, \ldots, h_N, J_{1,2}, \ldots, J_{N-1,N})$, the maximum probability of landing in an undesired state will be extremely close to one. To assist in distinguishing the probabilities, we minimize the maximum of the $\log$ of the probability of an undesired state:
        \begin{align*}
            \rho(\textbf{a}^{(1)}, \ldots, \textbf{a}^{(\ell)}) = 1 - \exp \left[ \min_{\psi} \max_{1\leq i \leq \ell} \log \mathbf{\text{Prob}_{boltz}}(\textbf{s}^{(i)},W^{(i)}) \right].
        \end{align*}
    This has the effect of amplifying small changes to the probability resulting from updates to the Hamiltonian coefficients.    
    
    An additional transformation is applied to the maximum function. 
    For a fixed set of Hamiltonian coefficients $\psi$, the maximum function 
    \begin{equation} \label{maxfunc}
    \max_i \left[ \log \mathbf{\text{Prob}_{boltz}} \left( \textbf{s}^{(i)}, W^{(i)} \right) \right] = \max_{i} \log \left[ \frac{\sum_{w\in W^{(i)}} \exp (H_\psi (w)) } {\exp(H_\psi(\textbf{s}^{(i)})) + \sum_{w\in W^{(i)}} \exp (H_\psi (w)) } \right] 
    \end{equation}
    is not guaranteed to be continuously differentiable.
    Because SLSQP requires the gradient to be defined at every point, we thus must approximate the maximum function. 
    For a sufficiently large scaling parameter $\lambda$ the following approximation holds for some arbitrary function $g$ evaluated at points $x_i$:
    \begin{equation} \label{logsum}
    \max_i \, g(x_i) \approx \frac{1}{\lambda} \log \sum_i \exp \{ \lambda \, g(x_i) \} \;\;\;\; \mbox{for} \;\;\;\; \lambda \gg 1
    \end{equation}
    Applying this approximation, we obtain a continuously differentiable version of equation (\ref{maxfunc}):
    \begin{align} \label{eq: approxmax}
    \max_i \, \left[ \log \mathbf{\text{Prob}_{boltz}}(\textbf{s}^{(i)},W^{(i)}) \right] \approx \frac{1}{\lambda} \log \sum_i \exp \left[ \lambda \, \log \mathbf{\text{Prob}_{boltz}}(\textbf{s}^{(i)},W^{(i)}) \right] 
    \end{align} 

   This completes our transformation process. The Boltzmann probability optimization function can now be approximated in a numerically stable, continuously differentiable form as follows:
            \begin{equation} \label{eq:datarho}
            \rho(\textbf{a}^{(1)}, \ldots, \textbf{a}^{(\ell)}) \approx 1 - \exp \left( \min_\psi f(\psi) \right)
        \end{equation}
    where, 
        \begin{equation} \label{logProbWrong}
         f(\psi) = \frac{1}{\lambda} \log \sum_i \exp \left[ \lambda \, \log \mathbf{\text{Prob}_{boltz}}(\textbf{s}^{(i)},W^{(i)}) \right]. 
        \end{equation}
    To speed up the work done by the SLSQP solver and to further ensure numerical stability, we also apply standard log-sum-exp transformations and vectorize $f(\psi)$. Details are found in the appendix. %\ref{gradient_appendix}.  %

    Our training data is of the form 
        \[\left(\text{input}, \text{ target}\right)  = \left(\textbf{a},  \rho(\textbf{a})\right) \]
    where $0\leq \rho(\textbf{a}) \leq 1$ and $\textbf{a} = (\textbf{a}^{(1)}, \ldots, \textbf{a}^{(\ell)})$ is the auxiliary array, where $\textbf{a}^{(i)} \in \{\pm1\}^\alpha$ is the vector of $\alpha$ auxiliary spins (see Section \ref{RIModel}). 
    
    For a given $\textbf{a}$, we use the SLSQP solver to minimize $f(\psi)$.
    From this we calculate the associated target value using equation (\ref{eq:datarho}).
    We use this procedure to generate training data for the four problems listed in Table \ref{table_prob}.

    \begin{table}[h]
    \centering
    \begin{tabular}{|c|c|c|c|c|c|}
        \hline
        Problem     & Number of Spins  & Fixed Spins   & Auxiliary Spins  & Number  & Dynamic Range \\
            &  $N$  &  $n$   & $\alpha$  & of Desired States &  of $h$, $J$ \\
        \hline \hline
        1           &         9           &           4  & 1             & 16 & [-4,4] \\
        2           &         11           &           5 & 1            & 32 & [-64,64] \\
        3           &         14           &           6 & 2            & 128 & [-256,256] \\
        4           &         15           &           6 & 3            & 192 & [-256,256] \\      
        \hline
    \end{tabular} 
    \caption{We consider four problems in this paper, to which we refer by problem number or its tuple $(N,n,\alpha)$, where $N$ is the total number of spins in the system and $n$ is the number of fixed spins. Problem 1 represents our 2-bit by 2-bit multiplication circuit configuration. Similarly, Problems 2, 3 and 4 represent our 2-bit by 3-bit, 2-bit by 4-bit, and 3-bit by 3-bit multiplication circuit configurations, respectively. The dynamic range column describes the continuous set of values that Hamiltonian coefficients $h, J$, that is the elements of $\psi$, can take in the SLSQP solver. The magnitude of the $h, J$ values has an effect on the range of the output probabilities returned by equation (\ref{eq:datarho}).
    Values in this column were determined empirically to yield training data covering a wide range of output probabilities.}
 \label{table_prob}
    \end{table}

 %%%%%%%%%%%%%%%%%%% rfmodel %%%%%%%%%%%%%%%%%%%
\subsection{Random Forest Regression Model}

Using the data collected in Section \ref{SectionData}, we train a random forest regressor, a collection of decision trees trained in parallel (independent from one another) such that, given an input, the output is the value most common (or average) to all of the decision trees~\cite{brieman2023rf}. 
For problem size $(N,n,\alpha)$, the input to the model is an auxiliary array, a set of $\textbf{a}\in \{\pm 1\}^\alpha$, and our target is $\rho(\textbf{a})$ as dictated by equation (\ref{eq:datarho}).

%To train a model for problem size $(N,n,\alpha)$,  our input for the model is the set of auxiliary arrays $\textbf{a}\in \{\pm 1\}^\alpha$ as described in Section \ref{SectionData} and similarly, $\rho(\textbf{a})$ our target as dictated by equation (\ref{eq:datarho}).

\begin{figure}[h] 
\centering
\subfloat[One Tree of the Random Forest]{
  \includegraphics[width=0.5\linewidth]{"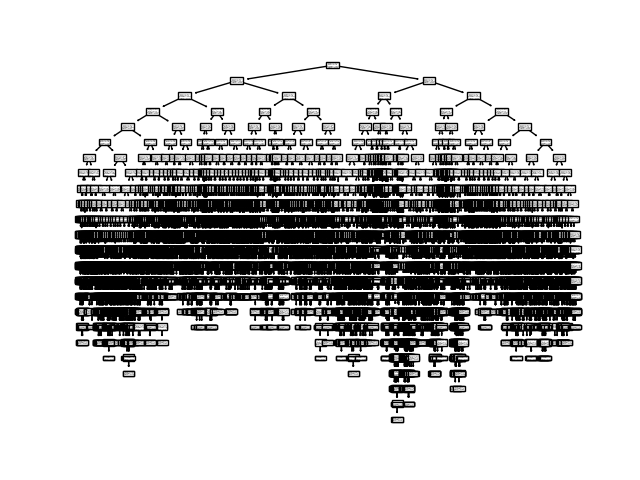"}
}
\subfloat[Set of Tree Nodes]{
  \includegraphics[width=0.5\linewidth]{"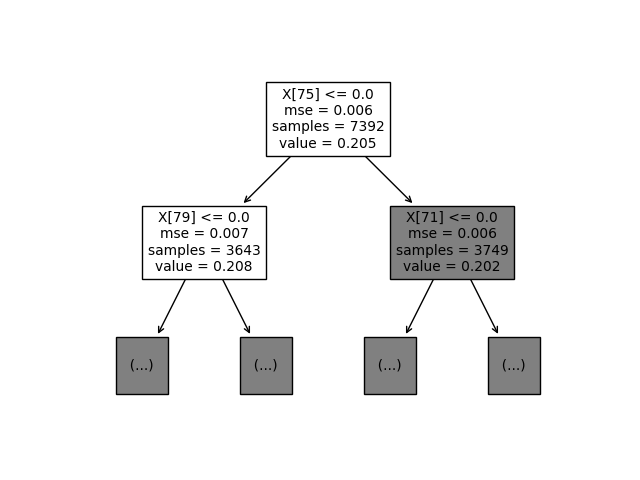"}
}
\caption{Subfigure (a) depicts one tree of a random forest created for Problem 3 (from Table \ref{table_prob}). Subfigure (b) shows an example set of nodes from the tree in subfigure (a). At each node of the tree, the choice of direction is  dependent on the value of a single element of the auxiliary array $\textbf{a}$. Note that the auxiliary array is denoted as $x$ in the figure. 
}
\label{Fig_RFTree}
\end{figure}

Due to the structure of random forests, the decisions to traverse the tree are discretized. That is, at each node of the tree, the choice of going left or right down the tree is 
dependent on the value of a single element of the auxiliary array $\textbf{a}$ (See Fig. \ref{Fig_RFTree}.) 

Despite this discreteness of the random forest, we obtain relatively good predictions as can be seen in Fig. \ref{BoltzRFFig}. Table \ref{table: rf} exhibits the mean squared error (MSE) of the estimated optimization function $\rho (\textbf{a})$ over the test data. For each problem, one hundred trees were used as an ensemble vote with a maximum tree depth of 27.

\begin{table}[h]
    \centering
    \begin{tabular}{|c|c|c|c|c|c|c|}
        \hline
        Problem & (N, n, $\alpha$) & Max Tree Depth
        & Size of Training Data & Size of Testing Data & MSE \\
        \hline \hline
        1 & (9 , 4 , 1) & 16 & 51773 & 12944 & 0.0001018 \\
        2 & (11 , 5 , 1) & 27 & 126428 & 35767 & 0.000473 \\
        3 & (14 , 6 , 2) & 16 & 7636 & 1910 & 0.019715 \\
        4 & (15 , 6 , 3) & 18 & 7477 & 1870 & 0.021173 \\
        \hline
    \end{tabular}
    \caption{The table provides the resulting Mean Squared Error (MSE) for our random forest regression model on the test data. Each regressor is described by the tree depth and size of the training data. Each problem pertains to a different size of auxiliary array $\textbf{a}$. As such, the amount of training and testing data changes and is provided for each problem. }
    \label{table: rf}
\end{table}

 %%%%%%%%%%%%%%%%%%% nnmodel %%%%%%%%%%%%%%%%%%%

\subsection{Deep Neural Network Model}

\begin{figure}[!h] 
\centering 
\subfloat[RF - Problem 1]{
  \includegraphics[width=0.24\linewidth,trim={2cm 0 1cm 0}]{"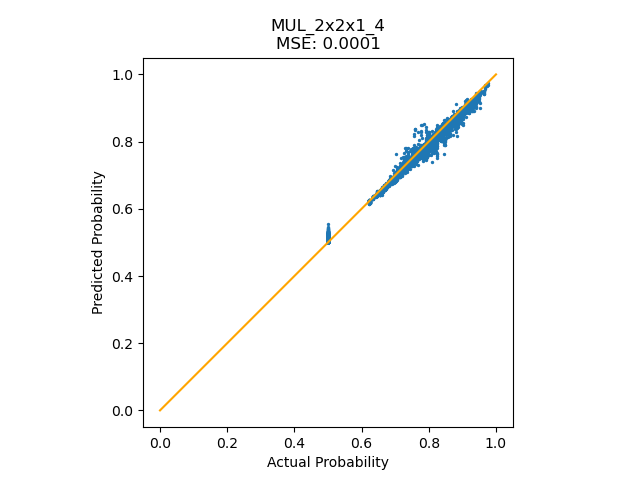"}
} 
\subfloat[RF - Problem 2]{
  \includegraphics[width=0.24\linewidth,trim={2cm 0 1cm 0}]{"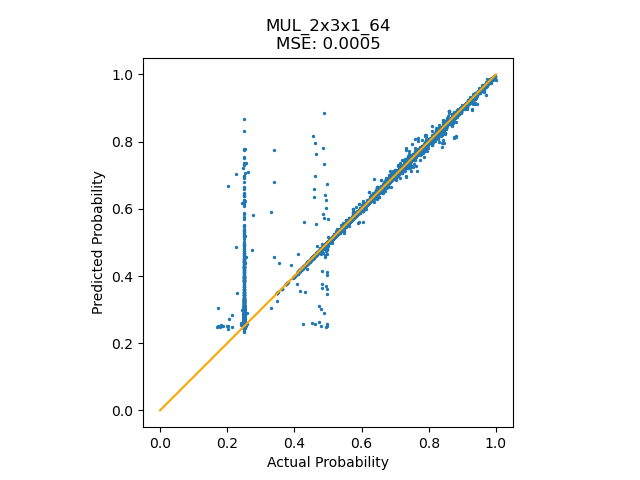"}
} 
\subfloat[RF - Problem 3]{
  \includegraphics[width=0.24\linewidth,trim={2cm 0 1cm 0}]{"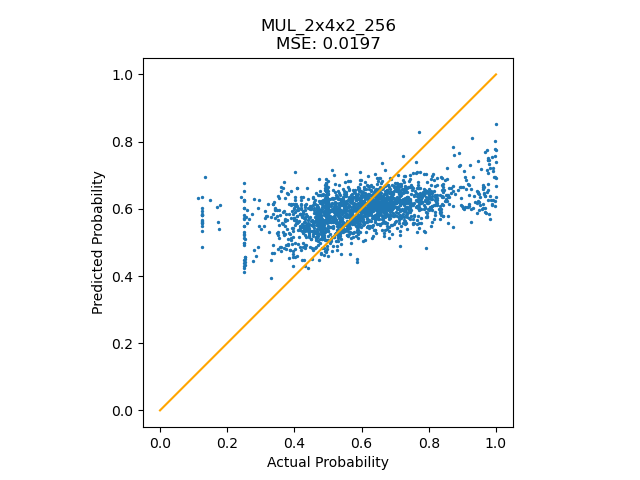"}
} 
\subfloat[RF - Problem 4]{
  \includegraphics[width=0.24\linewidth,trim={2cm 0 1cm 0}]{"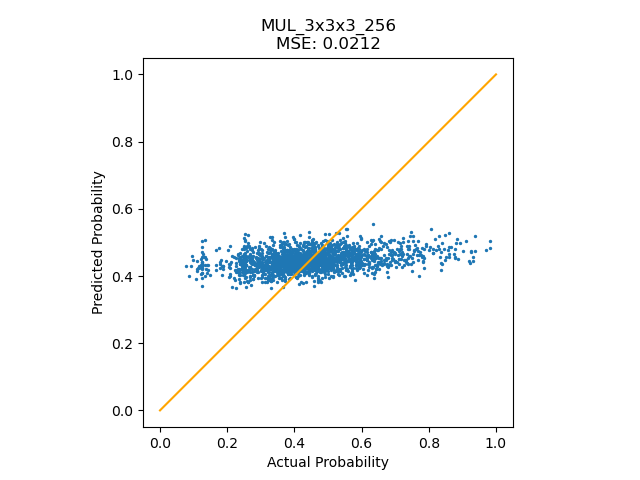"}
} \hspace{2mm}

\subfloat[DJINN- Problem 1]{
  \includegraphics[width=0.2\linewidth]{"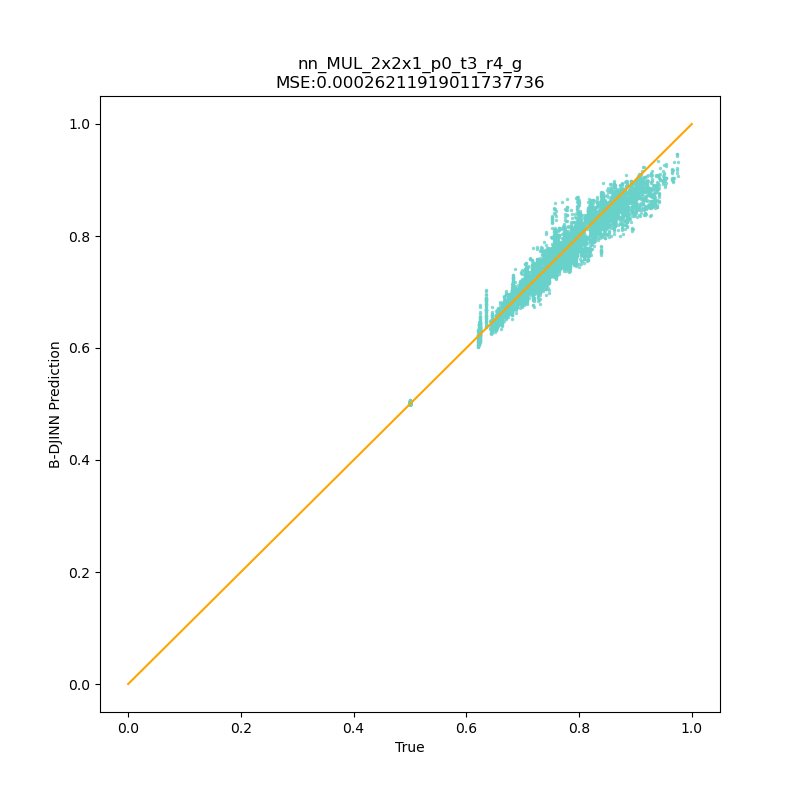"}
} \hspace{5mm}
\subfloat[DJINN - Problem 2]{
  \includegraphics[width=0.2\linewidth]{"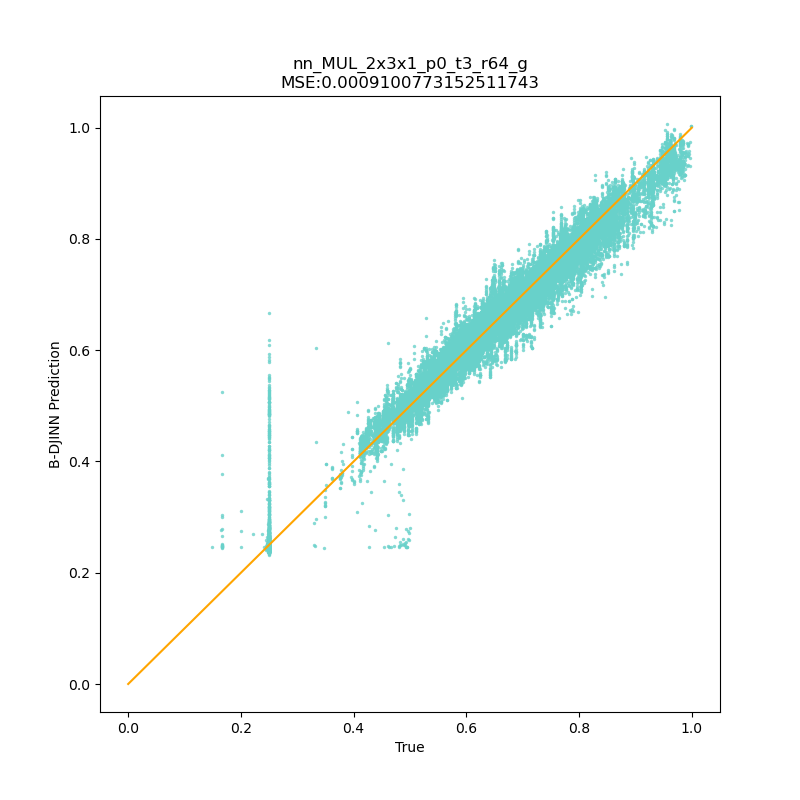"}
} \hspace{5mm}
\subfloat[DJINN - Problem 3]{
  \includegraphics[width=0.2\linewidth]{"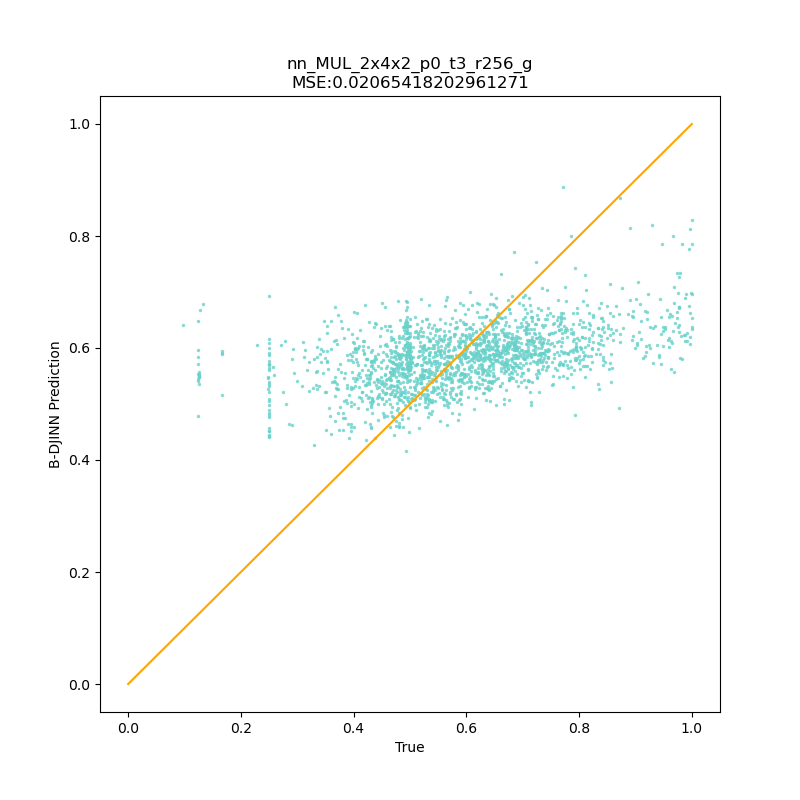"}
} \hspace{5mm}
\subfloat[DJINN - Problem 4]{
  \includegraphics[width=0.2\linewidth]{"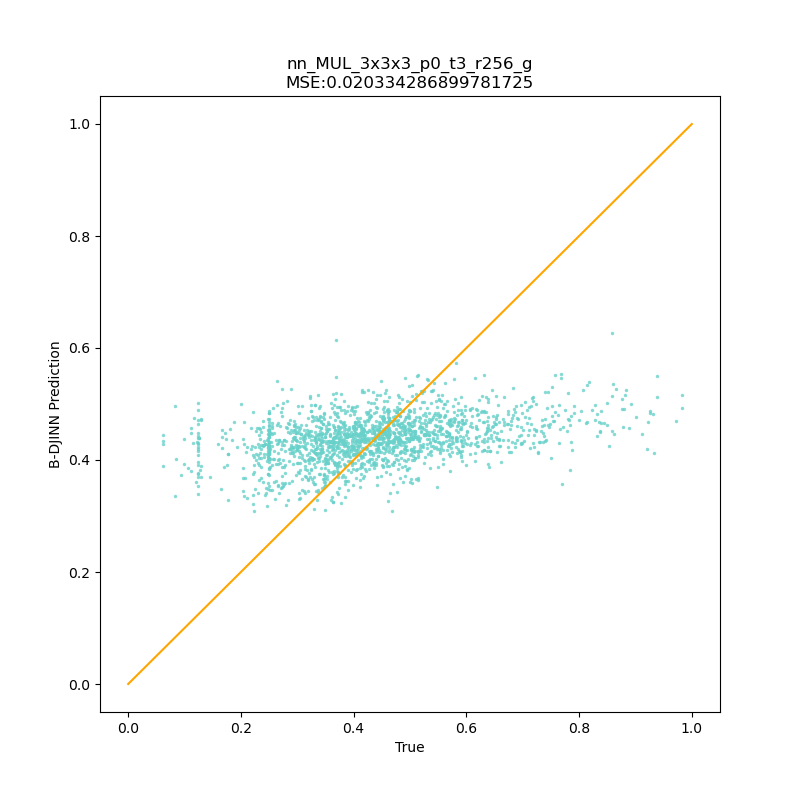"}
}

\caption{Subfigures (a)-(d) provide a visual representation of the results of the random forest regression models (denoted \emph{RF} in the figure) for the 4 problems listed in  Table~\ref{table: rf}. Subfigures (e)-(h) provide a visual representation of the results of DNN models created using DJINN framework (denoted \emph{DJINN} in the figure) for the 4 problems listed in Table~\ref{table: nn}. In each subfigure, the x-axis is the true Boltzmann probability, and the y-axis is the predicted probability. Each point represents the output corresponding to a single auxiliary array $\textbf{a}$.
}
\label{BoltzRFFig}
\end{figure}

In \cite{djinn}, Humbird et al. showed that a random forest can be modeled by a deep neural network (DNN). They provide a framework (DJINN) to go from a random forest to a DNN, using the random forest as an initialization of the DNN parameters. As our optimization function is non-linear and random forests are linear, we apply the DJINN framework to take our random forest regression model to a DNN, a non-linear model. Once the DNN parameters are initialized with the random forest, the model continues to learn and improves upon the initial random forest regressor. The MSE results can be seen in Table \ref{table: nn}  and Fig. \ref{BoltzRFFig}. 

The results are from DNNs grown from random forests with only three trees and a maximum tree depth of 10. For example, the neural network used  to model the data of Problem 1 has the following layer sizes:
\[ [  16,    18,    22,    30,    46,    78,   142,   270,   526, 1038,  2062,  4101,  8122, 15504, 24273, 27897,  1]  \]
Despite truncating the number of trees and the tree depth of the random forests used to initialize this DNN, one can see that the resulting neural network is able to model the data with similar MSE results to the random forest regression models. 
Furthermore, the random forest regressor in Problem 1 has 100 trees each at a depth of sixteen, which is approximately 65,535 nodes per tree. Consequently, the aggregated number of trainable parameters of the random forest regression models far exceeds that of the DNN models without significant improvement in prediction accuracy. 

\begin{table}[h]
    \centering
    \begin{tabular}{|c|c|c|c|c|c|c|}
        \hline
        Problem & (N, n, $\alpha$) 
        & Size of Training Data & Size of Testing Data & MSE \\
        \hline \hline
        1 & (9 , 4 , 1) &  51773 & 12944 & 0.000262  \\
        2 & (11 , 5 , 1) & 126428 & 35767 &  0.000910 \\
        3 & (14 , 6 , 2) & 7636 & 1910 & 0.0206542\\
        4 & (15 , 6 , 3) & 7477 & 1870 & 0.0203343 \\
        \hline
    \end{tabular}
    \caption{The table provides the resulting Mean Squared Error (MSE) for our deep neural network models. The neural networks, created using DJINN framework, were initialized with a random forest regressor with only three trees and a maximum depth of ten; whereas our random forest regressors had 100 trees with larger depth. Consequently,
    the number of parameters required is far fewer than that required for the random forest regression model and yet, the MSE of the results are similar.}
    \label{table: nn}
\end{table}

%%%%%%%%%%%%%%%%%%% Comparisons %%%%%%%%%%%%%%%%%%%

 \section{Performance Evaluations}
\label{comparisons}
        
   In Table ~\ref{table_compare}, we provide performance evaluations of four  on Problem 4 from Table ~\ref{table_prob}. 
   The first two  are implementations of the  SLSQP algorithm, the latter two are our ML models. 
   
   The default behavior of SciPy's SLSQP algorithm involves approximating the gradient numerically. 
   This version of the solver is referred to as ``SLSQP (Approx Grad)''. 
    For high-dimensional problems such as the reverse Ising problem, this approximation is computationally expensive because the number of objective evaluations required scales with the number of dimensions (spins).
    To mitigate this, we explicitly implemented the gradient in NumPy and pass it to the solver.   
    This version of the code is referred to as ``SLSQP (Explicit Grad)''. See the appendix for details.
    %See ~\ref{gradient_appendix} for details.

\begin{table}[h]
    \centering 
\begin{tabular}{|l||c|c|} 
    \hline
    Model         &    Time to compute 100 values    &   Average time to compute 1 value \\
    \hline \hline
    SLSQP (Approx Grad)          &   4 days, 5:01:24.4              &          60.6 mins                 \\
    \hline
    SLSQP (Explicit Grad)       &    5:08:53.7                     &          3.09 mins                 \\
    \hline
    Random Forest Regressor   &    31.987 ms after 258 secs training    &          320 microseconds                \\
    \hline
    DJINN DNN       &  28.5 ms after 251 secs training          &    285 microseconds                \\
    \hline
\end{tabular}
\caption{The performance comparison above considers four methods for computing Problem 4 in Table~\ref{table_prob}. It shows the time required to calculate the min-max Boltzmann probability for an ensemble of 100 computations starting from different auxiliary spins. Half of the auxiliary spins were randomly selected from values that satisfy the constraints of equation (\ref{eq_H0}) while the other half were selected from values that did not. The final column extrapolates the nominal average times to perform a single Boltzmann probability prediction for any randomly chosen set of initial auxiliary spins.The SLSQP solvers and random forest regressors were run in serial on an AMD EPYC 7713 64-Core processor running at 2 Ghz. The DNNs were run on a single NVDIA A6000 GPU. DJINN has an additional overhead in determining optimum number of parameters.}
\label{table_compare}
\end{table}

The DJINN framework enabled us to construct DNNs that predict optimal Boltzmann probabilities for the reverse Ising problem in significantly less time than the current state of the art. With an MSE of 0.02 or less, we conclude that it should be possible to construct acceptably accurate predictors of Ising system dynamics based on deep learning models trained with the output of our Boltzmann probability solver. Those trained models can be used to analyze much larger spin configurations than is currently possible in order to design better ground state solutions for the reverse Ising problem.

%%%%%%%%%%%%%%%%%%%%% inclue references

\bibliographystyle{is-plain}
\bibliography{refs}

%%%%%%%%%%%%%%%%%%%%% online methods section 
\newpage
\section*{Appendix}
\label{gradient_appendix}

%\subsection{Efficient Gradient Implementation}
\subsection*{Computing the Gradient}
    For the system being considered in the reverse Ising problem, all possible states can be divided into two sets. The first set consists of desired, or ``correct'' states, and the remaining states constitute the undesired or ``incorrect'' states.
    \begin{equation}
    \mathbf{r}^\top = \exp(H(s^{(i)})) = 
    \mathbf{1}^\top \cdot \exp \{ - \beta \, \mathbf{R} \cdot \boldsymbol{\psi} \}
    \end{equation}
    \begin{equation}
    \mathbf{w}^\top = \sum_{w\in W^{(i)}} \exp (H(w)) =
    \mathbf{1}^\top \cdot \exp \{ - \beta \, \mathbf{W} \cdot \boldsymbol{\psi} \}
    \end{equation}
    Using this notation, the objective function can be written as the sum-exp of the log of the maximum probability of the Ising system evolving to an undesirable state
    \begin{equation}
    f(\boldsymbol{\psi}) = \frac{1}{\lambda} \left( \mathbf{1}^\top \cdot \exp \{ \lambda \log \mathbf{p}_W \} \right) \;\;\;\; \mbox{where} \;\;\;\;  \mathbf{p}_W = \frac{\sum_{w\in W^{(i)}} \exp (H(w)) } {\exp(H(s^{(i)})) + \sum_{w\in W^{(i)}} \exp (H(w)) }
    \end{equation}
    Consider the partial of the sum-exp's associated with the Hamiltonians of the correct and incorrect states. These can be neatly expressed as the weighted sum-exp of the exponential terms using the element-wise multiply (i.e., broadcast) denoted by $\odot$.
    \begin{equation}
    \frac{\partial \mathbf{r}}{\partial \boldsymbol{\psi}}^\top = - \beta \, \mathbf{1}^\top \cdot \big( \exp \{ - \beta \, \mathbf{R} \cdot \boldsymbol{\psi} \} \odot \mathbf{R} \big)
    \end{equation}
    \begin{equation}
    \frac{\partial \mathbf{w}}{\partial \boldsymbol{\psi}}^\top = - \beta \, \mathbf{1}^\top \! \cdot \big( \exp \{ - \beta \, \mathbf{W} \cdot \boldsymbol{\psi} \} \odot \mathbf{W} \big)
    \end{equation}
    The gradient of the log of the probability of an incorrect state for an input level is given in terms of the probabilities of correct states below.
    \begin{equation}
    \frac{\partial \log \mathbf{p}_W}{\partial \boldsymbol{\psi}} = \frac{\mathbf{p}_R}{\mathbf{r}}^\top \!\! \odot \frac{\partial \mathbf{r}}{\partial \boldsymbol{\psi}} - \frac{\mathbf{p}_R}{\mathbf{w}}^\top \!\! \odot \frac{\partial \mathbf{w}}{\partial \boldsymbol{\psi}}
    \end{equation}
    where
    \begin{equation}
    \mathbf{p}_R^\top = 1 - \mathbf{p}_W^\top
    \end{equation}
    And thus, the gradient of the objective function that is an approximation of the maximum probability of an incorrect output is given below.
    \begin{equation}
    \frac{\partial f}{\partial \boldsymbol{\psi}} = \frac{\partial \log \mathbf{p}_W}{\partial \boldsymbol{\psi}} \cdot \frac{ \exp \{ \lambda \log \mathbf{p}_W \} }{ \mathbf{1}^\top \cdot \exp \{ \lambda \log \mathbf{p}_W \} }
    \end{equation}

\subsection*{Gradient Implementation}
    Notice that the gradient can now be expressed in terms of fast, dense linear algebra primitives. The code below demonstrates the Boltzmann objective $f(\psi)$ and gradient $\partial f / \partial \psi$ vectorized and implemented in pure Python for up to 20-fold improvement in computational efficiency while preserving accuracy.

\begin{lstlisting}[language=Python]    
import numpy as np

def boltzmannObjective(beta, large, R, W, psi):

    _, _, _, lseW, lseZ = _boltzmannUtils(beta, R, W, psi)

    logPrW = lseW - lseZ
    logPrWMax = _boltzmannLogSumExp(large * logPrW) / large

    return logPrWMax

def boltzmannGradient(beta, large, R, W, psi):

    r, w, lseR, lseW, lseZ = _boltzmannUtils(beta, R, W, psi)

    logPrW = lseW - lseZ
    logMaxPrW = _boltzmannLogSumExp(large * logPrW) / large

    dsoR = _boltzmannSumExpOffset(lseR, R.T, -beta * r)
    dsoW = _boltzmannSumExpOffset(lseW, W.T, -beta * w)
    PrR, dLogPrW = 1 - np.exp(logPrW), logPrW.T - logMaxPrW

    return ( beta * PrR * ( dsoR - dsoW ) ) @ np.exp( large * dLogPrW )

def _boltzmannLogSumExp(M):

    x = np.amax(M, axis=0)
    return x + np.log( np.sum( np.exp(M - x), axis=0 ) )

def _boltzmannSumExpOffset(y, T, M):

    x = np.amax(M, axis=0)
    return np.sum( np.exp(M - x).T * T, axis=2 ) / np.exp(y - x)

def _boltzmannUtils(beta, R, W, psi):

    r, w = R @ psi, W @ psi
    z = np.concatenate((r, w))

    lseR = _boltzmannLogSumExp(-beta * r)
    lseW = _boltzmannLogSumExp(-beta * w)
    lseZ = _boltzmannLogSumExp(-beta * z)

    return r, w, lseR, lseW, lseZ

\end{lstlisting}

\end{document}